\documentclass[letterpaper, 10 pt, conference]{ieeeconf}  

\IEEEoverridecommandlockouts                              

\makeatletter
\let\NAT@parse\undefined
\makeatother
\usepackage[pagebackref=false,breaklinks=true,colorlinks=true,bookmarks=false,linkcolor={red!50!black},urlcolor={blue},citecolor={green!60!black}]{hyperref} 
\usepackage{url}
\usepackage{pifont}
\usepackage{booktabs} 

\usepackage{etoolbox}
\makeatletter
\patchcmd{\@makecaption}
  {\scshape}
  {}
  {}
  {}
\makeatletter
\patchcmd{\@makecaption}
  {\\}
  {.\ }
  {}
  {}
\makeatother

\usepackage{graphicx}
\usepackage{subfigure}
\usepackage{lipsum}
\usepackage{cite}
\usepackage{bm}
\usepackage{esvect}
\usepackage{amsmath}
\usepackage[ruled,vlined,lined,boxed,commentsnumbered]{algorithm2e}
\usepackage{color}
\usepackage{amsmath,amssymb}
\usepackage{tabularx}
\usepackage{multirow}
\usepackage{arydshln}
\usepackage[dvipsnames]{xcolor}

\usepackage{overpic}
\usepackage{overpic} 
\usepackage{color}
\usepackage{mathtools} 
\usepackage{currfile}
\usepackage{multirow}

\definecolor{turquoise}{cmyk}{0.65,0,0.1,0.3}
\definecolor{purple}{rgb}{0.65,0,0.65}
\definecolor{dark_green}{rgb}{0, 0.5, 0}
\definecolor{orange}{rgb}{0.8, 0.6, 0.2}
\definecolor{red}{rgb}{0.8, 0.2, 0.2}
\definecolor{darkred}{rgb}{0.6, 0.1, 0.05}
\definecolor{blueish}{rgb}{0.0, 0.3, .6}
\definecolor{light_gray}{rgb}{0.7, 0.7, .7}
\definecolor{pink}{rgb}{1, 0, 1}
\definecolor{greyblue}{rgb}{0.25, 0.25, 1}

\renewcommand{\paragraph}[1]{\vspace{.5em}\noindent\textbf{#1}.}

\usepackage{lipsum}

\usepackage{blindtext}

\newcommand{\kostas}[1]{{\color{Bittersweet} {[\bf Kosta: #1]}}}
\newcommand{\andrew}[1]{{\color{BlueViolet} {[Andrew: #1]}}}
\newcommand{\vitto}[1]{{\color{OrangeRed} {[Vitto: #1]}}}
\newcommand{\JB}[1]{{\color{OliveGreen} {[Jon: #1]}}}
\newcommand{\tom}[1]{{\color{RoyalPurple} {[Tom: #1]}}}
\newcommand{\pratul}[1]{{\color{Emerald} {[Pratul: #1]}}}
\newcommand{\at}[1]{{\color{blueish}#1}}
\newcommand{\AT}[1]{{\color{blueish}{\bf [Andrea: #1]}}}
\newcommand{\At}[1]{\marginpar{\tiny{\textcolor{blueish}{#1}}}}
\newcommand{\al}[1]{\textbf{\color{orange}[AL: #1]}}
\renewcommand{\kostas}[1]{}
\renewcommand{\andrew}[1]{}
\renewcommand{\vitto}[1]{}
\renewcommand{\JB}[1]{}
\renewcommand{\tom}[1]{}
\renewcommand{\pratul}[1]{}
\renewcommand{\at}[1]{}
\renewcommand{\AT}[1]{}
\renewcommand{\At}[1]{}
\renewcommand{\al}[1]{}

\usepackage[lined,boxed,commentsnumbered]{algorithm2e}

\makeatletter
\usepackage{xspace}
\DeclareRobustCommand\onedot{\futurelet\@let@token\@onedot}
\def\@onedot{\ifx\@let@token.\else.\null\fi\xspace}

\makeatother

\definecolor{gold}{rgb}{0.82, 0.53, 0.04}
\definecolor{silver}{rgb}{0.43, 0.47, 0.35}
\definecolor{bronze}{rgb}{0.5, 0.51, 0.78}

\allowdisplaybreaks
\title{\LARGE \bf
Robust Multimodal Dynamic Object Segmentation
}

\author{Zhe Xin$^{1*\dagger}$, Hanzhi Chang$^{1*}$,  Penghui Huang$^{1}$,  Yinian Mao$^{1}$,
Guoquan Huang$^{1,2}$ 
\thanks{$^{1}$ Meituan UAV, Beijing, China.
        {\tt\small \{xinzhe, changhanzhi, huangpenghui03, maoyinian\}@meituan.com}}
\thanks{$^{2}$ Dept. of Mechanical Engineering, Computer and Information Sciences, University of Delaware, Newark, DE, USA. {\tt\small ghuang@udel.edu}}%
\thanks{\tt\small $\dagger$ denotes corresponding author.}
\thanks{\tt\small$*$ Authors contributed equally to this work.}
}

\begin{document}

\maketitle
\thispagestyle{empty}
\pagestyle{empty}

\begin{abstract}
Dynamic object segmentation plays a critical role in many visual applications such as static scene reconstruction from dynamic videos. However, existing optical flow-based methods fail to ensure consistent static/dynamic segmentation along object boundaries, while 3D reconstruction-based approaches are highly sensitive to reconstruction errors. To address these limitations, we present a dynamic object segmentation framework that can generate both precise and complete dynamic masks by integrating multimodal cues including 2D point tracks, 3D reconstruction, and semantic information. 
We design a network combining Transformer architectures with feature clustering aggregation modules to perform static/dynamic classification of multimodal feature trajectories. It enables the model to adaptively determine which type of feature should dominate based on the characteristics of each scene, while also mitigating the impact of feature degradation. Additionally, we introduce a novel point-query-based SAM post-processing method capable of handling multiple objects within a single mask. Extensive experiments demonstrate that our approach achieves state-of-the-art performance in both dynamic object segmentation and static scene reconstruction tasks.
\end{abstract}

\section{Introduction}

Dynamic object segmentation is essential in various domains including autonomous robotics, AR/VR, and video understanding~\cite{hammer2016motion, klappstein2009moving, rashed2019motion, wang2020dymslam}. 
This technique plays a crucial role in reconstructing static scenes from dynamic videos, as it enables the automatic filtering of moving objects while retaining static scene elements essential for 3D reconstruction. However, in real-world settings, due to complex object and camera motion patterns, environmental interferences, and variations in the scale of dynamic objects, dynamic object segmentation remains a challenging task.


Current work on dynamic object segmentation can be categorized into two paradigms: 2D optical flow based approaches and 3D-reconstruction-based techniques.
2D-based methods~\cite{zhao2022particlesfm, chen2024leap, yang2021self, zhou2020matnet} typically estimate the optical flow from dynamic video sequences and perform static/dynamic classification at the pixel level.
3D reconstruction-based methods~\cite{zhang2024monst3r, chen2025easi3r, xu2024das3r} first estimate depth maps or 3D point clouds from dynamic scenes, and subsequently derive dynamic masks through post-processing. The ViT architectures used in these methods can inherently capture semantic information, improving mask-object boundary alignment. 

\begin{figure}
    \centering
    \includegraphics[width=0.90\linewidth]{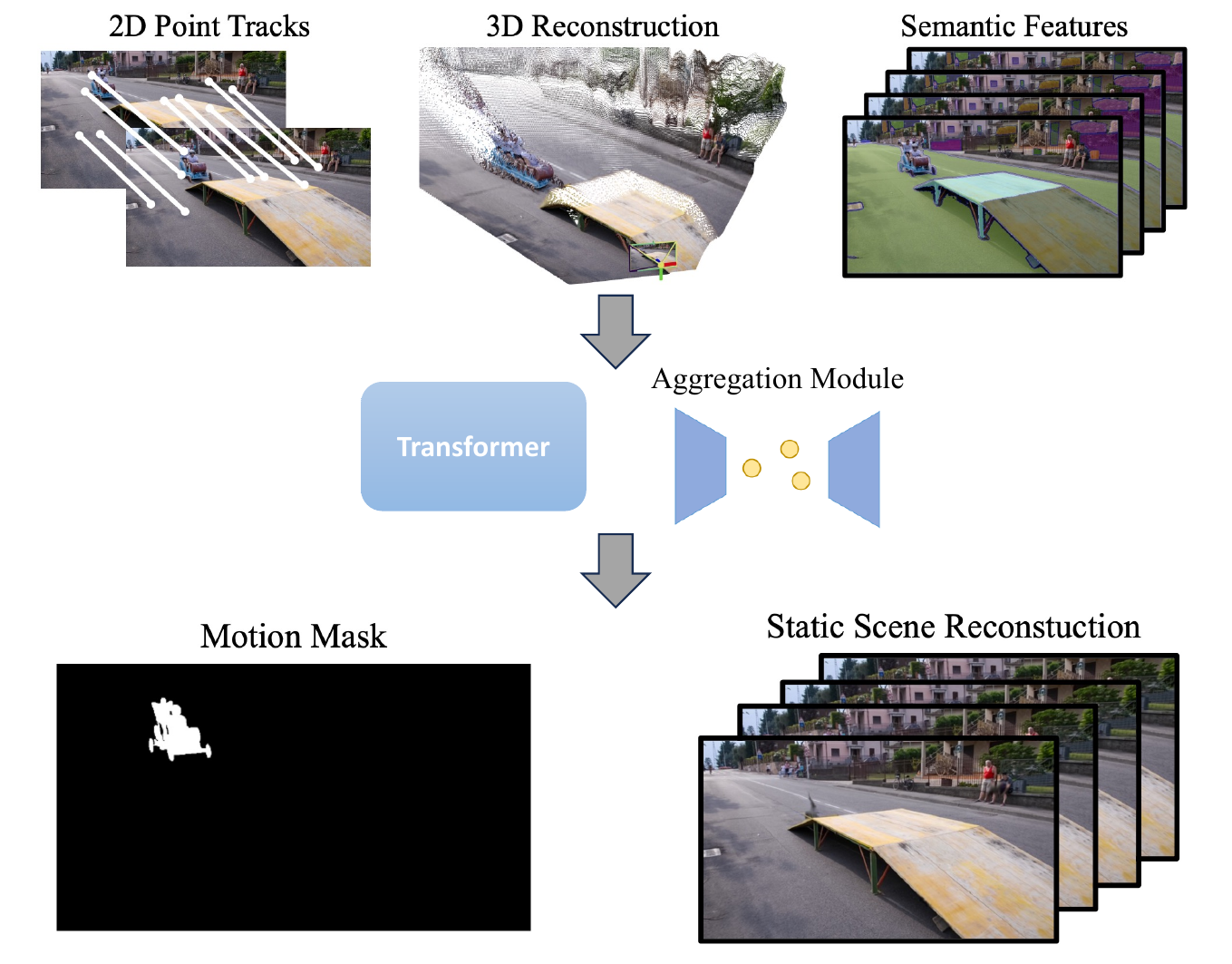}
    \caption{\textbf{Overview.} Given a dynamic video containing several image frames, Our method leverages 2D point tracking, 3D reconstruction, and semantic priors as inputs, to accomplish the tasks of dynamic object segmentation and static scene reconstruction from dynamic videos.}
    \label{fig:placeholder}
\end{figure}

In order to generate both precise and complete dynamic object masks, relying solely on a single modality cue is often insufficient. 
For example, \textit{2D optical flow or point tracking} directly reflects pixel-level movements that are intrinsically correlated with object motions. However, their pixel-wise nature fails to ensure consistent static/dynamic segmentation within object boundaries, and the lack of higher-dimensional information prevents the decoupling of camera and object motions.
\textit{3D reconstruction} provides approximate camera poses and 3D point map distributions, offering higher-dimensional perspective information for dynamic object segmentation. Nevertheless, its dynamic masks are highly sensitive to errors in depth estimation and 3D point map generation. Additionally, dynamic objects frequently violate epipolar constraints, leading to particularly unreliable 3D estimations for moving targets in certain scenarios.
\textit{Semantic information} serves as a valuable reference for maintaining segmentation consistency within objects. Current state-of-the-art methods often require SAM-based post-processing. However, they typically feed the entire mask as a single query to SAM, assuming all masked pixels belong to a single object, which is not always the case in real-world scenarios. 

To address the limitation of single-modal cues, in this paper, we design a unified network through the integration of optical flow, 3D reconstruction, and semantic information. Our approach takes semantic feature maps from SAM, 3D reconstruction results including depth maps, camera intrinsic/extrinsic parameters, and attention maps from MonST3R and 2D point tracks between multiviews as input, using feature trajectory classification paradigm to perform dynamic object segmentation, which includes a Transformer for multi-frame and multi-track information integration and a point-based aggregation network for feature clustering. To enhance mask quality in multi-object scenarios, we develop a SAM-enhanced post-processing technique that utilizes individual pixels rather than entire masks as SAM queries and iteratively refines initial masks. Experimental results demonstrate state-of-the-art performance in dynamic object segmentation. Furthermore, to validate the effectiveness of our method, we combine mask generation, pose estimation, and 3D Gaussian Splatting (3DGS) reconstruction to perform static scene reconstruction from dynamic videos. Experiments show superior performance compared to existing SOTA methods.

In summary, the main contributions of this paper include:
\begin{itemize}
\item We propose a dynamic object segmentation framework that integrates multimodal cues, including 2D point tracking, 3D reconstruction, and semantic information to achieve superior segmentation performance.
\item We introduce a SAM-based post-processing methodology to address complex real-world dynamic scenarios involving multiple moving objects, significantly improving mask-based post-processing approaches.
\item We conduct extensive validations demonstrating the state-of-the-art performance of the proposed method across multiple benchmarks in both the accuracy of motion masks and the quality of static scene reconstruction.
\end{itemize}

\section{Related Work}
\label{sec:related_work}

Dynamic object segmentation focuses on predicting motion masks from video inputs. Traditional structure-from-motion methods, such as COLMAP, operate under the assumption that scenes are predominantly static, making them ineffective for videos containing dynamic objects. Alternative approaches~\cite{bescos2018dynaslam,xiao2019dynamic,ungermann2024robust} prioritize tracking and semantically segmenting dynamic objects but often rely on assumptions, such as distinguishing foreground object motion from a static background or pre-identifying mobile objects. These constraints limit their general applicability.

Classical methods typically employ optical flow estimation~\cite{lian2023bootstrapping, xie2022segmenting, yang2021self} and point tracking~\cite{karazija2024learning, brox2010object} to separate moving objects from the background. To address these challenges, CasualSAM~\cite{zhang2022structure} jointly optimizes depth (using a pre-trained learned prior), camera poses, and motion masks. ParticleSfM~\cite{zhao2022particlesfm} utilizes off-the-shelf optical flow and monocular depth estimators to create 3D tracks and trains a 3D motion classifier on synthetic data. Similarly, LEAP-VO~\cite{chen2024leap} classifies tracks into static and dynamic components, enhances inputs with additional features, employs a refiner module, and uses a sliding window approach for global bundle adjustment to estimate camera poses. RoMo~\cite{goli2024romo} incorporates unreliable epipolar geometry through Sampson error~\cite{sampson1982fitting} and SAMv2~\cite{ravi2024sam2}. However, as these methods are trained exclusively on 2D data, they often struggle with issues such as imprecise optical flow, occlusions, and distinguishing object motion from camera motion, particularly in scenarios with faulty correspondences.

For point-map-based methods, DUSt3R~\cite{wang2024dust3r} introduces a novel pose inference technique using a patch-based feed-forward network to predict global 3D coordinates. MonST3R~\cite{zhang2024monst3r} fine-tunes DUSt3R for dynamic scenes, integrating optical flow~\cite{wang2024sea} with estimated pose and depth to achieve dynamic object segmentation. Building on this, DAS3R~\cite{xu2024das3r} trains a DPT~\cite{ranftl2021vision} on top of MonST3R to enable feed-forward segmentation estimation. While effective, these approaches heavily rely on accurate pose and depth estimation to maintain reprojection consistency and require extensive training on diverse motion patterns to achieve robust generalization. Easi3R exploits attention layers from pre-trained MonST3R models to extract dynamic segmentation, it still cannot generalize to situations where MonST3R performs poorly, and fails to recover mask details without SAM-based post-processing due to the low resolution of attention maps. 

\section{Dynamic Object Segmentation with Multimodal Cues} 
\label{sec:method}

\begin{figure*}
    \centering
    \includegraphics[width=0.95\linewidth]{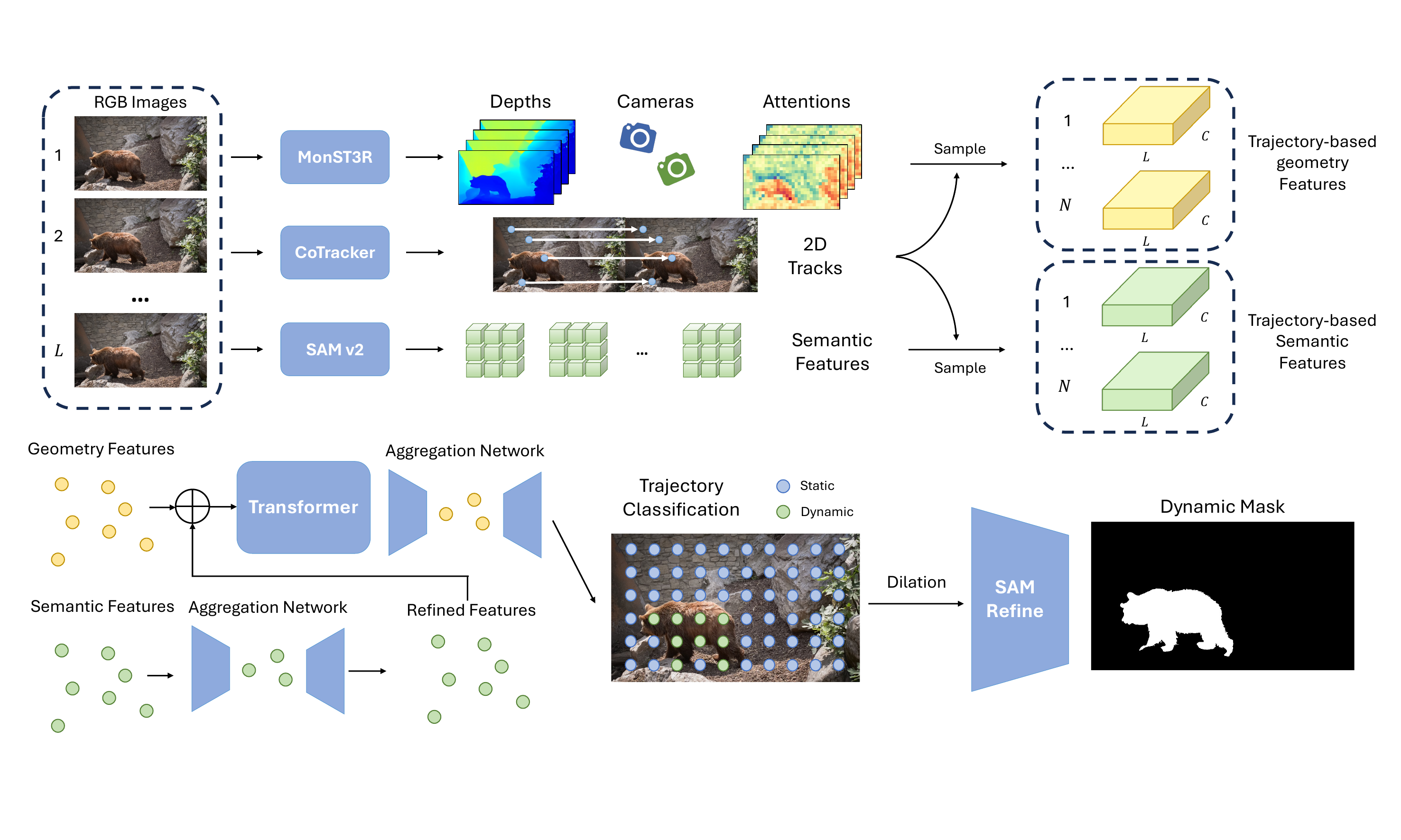}
    \vspace{-10mm}
    \caption{\textbf{Overall Architecture.} Given a video clip with $N$ frames, we first employ MonST3R~\cite{zhang2024monst3r} to perform coarse 3D reconstruction of the dynamic scene, obtaining depth information, camera parameters, and attention maps. Simultaneously, we use CoTracker~\cite{karaev24cotracker3} to densely sample 2D points on the images and obtain their motion trajectories across frames. These 2D point trajectories are then associated with corresponding 3D reconstruction features and semantic features are extracted from SAM~\cite{ravi2024sam2} feature maps. These combined features are fed into our network consisting of a Transformer module and two aggregation networks. This network ultimately outputs static/dynamic classification results for each feature trajectory. To obtain the final high-quality dynamic masks, we apply our proposed SAM-based post-processing method to refine the initial dynamic masks.}
    \label{fig:placeholder}
    \vspace{-3mm}
\end{figure*}

Given a dynamic video clip containing $N$ frames $\{I_i\}_{i=1}^N$, our goal is to estimate dynamic masks $\{M_i\}_{i=1}^N$ for each frame. 
Fig.~\ref{fig:placeholder} illustrates the overall
system architecture. The primary components include multimodal feature extraction, dynamic mask estimation, and SAM-based post-processing. 
In the following, we will explain these modules in detail.

\subsection{Multimodal Features Extraction}
\label{sec:preprocess}

The framework effectively integrates 2D tracking, 3D reconstruction, and semantic information to achieve accurate dynamic mask predictions. 

For 3D reconstruction, we employ MonST3R~\cite{zhang2024monst3r}, which processes the entire video clip using its ViT architecture with global alignment to generate per-frame depth maps $\{D_i\}_{i=1}^N$, camera intrinsic parameters $\{K_i\}_{i=1}^N$, and camera extrinsic parameters $\{P_i | R_i, t_i\}_{i=1}^N$. Building upon insights from Easi3R~\cite{chen2025easi3r} that the attention maps from the ViT decoder contain valuable information for identifying dynamic objects, we also compute and aggregate per-frame attention maps $\{A_i\}_{i=1}^N$. These features are then used in the feature trajectory construction process.

By extracting the self-attention maps and cross-attention maps from the ViT decoder layer, through averaging operations performed along query and layer dimensions, the corresponding attention maps are obtained for each image pair. Subsequently, we compute the mean and variance of attention maps across these pairs, and concatenate the results to generate the final attention map. 

For semantic information, we utilize SAMv2~\cite{ravi2024sam2} to extract per-frame semantic feature maps $\{F_i\}_{i=1}^N$. For 2D tracking information, we sample $200\times200$ 2D points from the first video frame and use them as queries into CoTracker~\cite{karaev24cotracker3} to obtain 2D tracking results $\{x_i^k\}_{i=1}^N$ for each pixel $k$.

\subsection{Dynamic Mask Estimation}
\label{sec:classification}
The multimodal features extracted in the preceding section  are first converted into feature trajectories, which are then fed into the main model to classify each trajectory. These results are subsequently transformed into coarse dynamic masks using image dilation and refined by the SAM point-query-based iterative strategy.

\subsubsection{Formulate Features Trajectories}

We first convert input features into point-based trajectories, as this representation is more general and effectively captures information across the entire image sequence compared to two-frame flow models. For the $k$-th track $\{x_i^k\}_{i=1}^N$, we obtain per-frame depth values $\{d_i^k\}_{i=1}^N$ and attention values $\{a_i^k\}_{i=1}^N$ through interpolation, we then compute pixel position offsets $\Delta x_i^k$ and depth offsets $\Delta d_i^k$ between consecutive frames. Incorporating offset information provides a better representation of the changes between frames. Additionally, pixel position, depth, and offset are highly decoupled from the image content, which enhances the generalization capability of our method.
\begin{equation}
    d_i^k,a_i^k=\mathrm{Interpolation}{(D_i,A_i)}
\end{equation}
\begin{equation}
    \Delta x^k_i=x^k_{i+1} - x^k_i,\ \ \ \ \Delta d^k_i=d^k_{i+1} - d^k_i
\end{equation}

Moreover, we employ camera parameters, including the focal length $f_i=(f^x_i,f^y_i)$ derived from intrinsic $\{K_i\}_{i=1}^N$ and extrinsics $\{P_i\}_{i=1}^N$. The transformation matrices $\{\Delta P_i | \Delta R_i, \Delta t_i\}_{i=1}^N$ between adjacent frames are also computed. Although camera information is inherently embedded in optical flow and depth, we find that explicitly providing these features as input improves performance, since this simplifies the learning difficulty for the model and enables more direct and effective utilization of camera motion. 


The aforementioned features, such as 2D optical flow, depth, and camera poses, primarily describe the relationships of individual trajectories across images but fail to capture the consistency across multiple trajectories. Ignoring this consistency may lead to incomplete dynamic object segmentation. To address this limitation, we incorporate semantic features $c^k_i$ obtained by interpolating the semantic feature maps. By combining coarse-grained attention features with fine-grained semantic features, our method achieves more accurate and complete dynamic masks. The final feature trajectories are formulated as follows, where $||$ means concatenation,

\begin{equation}
    \mathbf F_\mathrm{traj}^k=(x^k\ ||\ \Delta x^k\ ||\ d^k\ ||\ \Delta d^k\ ||\ c^k\ ||\ a^k)
\end{equation}
\begin{equation}
    \mathbf F_\mathrm{sem}^k=\mathrm{Sample}(\{F_i\}_{i=1}^N)
\end{equation}



\subsubsection{Trajectory Classification}
The semantic features of pixels belonging to the same object exhibit similar characteristics. To better capture pixel-wise relationships, we first utilize a point-based module $\Phi_{\mathrm{sem}}$ to process and cluster semantic features $\mathbf F_\mathrm{sem}^{k}$.
This module clusters multiple feature trajectories using differentiable pooling operations and employs CNNs to facilitate interaction, thereby capturing global information in each cluster. The global information then aggregates with each local features in the cluster through unpooling operations to obtain the local-global context-aware features $\mathbf F_\mathrm{sem}^{k'}$. This aggregation network unifies similar semantic features through feature clustering, providing meaningful semantic priors for subsequent processing.

Next, we concatenate $\mathbf F_\mathrm{traj}^{k}$ with $\mathbf F_\mathrm{sem}^{k'}$ and input them into a Transformer encoder to enable interaction in each trajectory and among all trajectories. These two types of features are processed separately because $\mathbf F_\mathrm{traj}^{k}$ primarily describes geometric properties with minimal correlation to image content, whereas $\mathbf F_\mathrm{sem}^{k'}$ encodes texture attributes that are closely tied to image content. 

Then, the refined features are fed into an aggregation network $\Phi$ acting as the decoder, where they are further clustered and fused. This process generates motion confidence scores for each trajectory, with the final classification determined via a sigmoid function. The entire procedure can be formally expressed as:
\begin{equation}
m^k = \mathrm{Sigmoid}\left(\Phi\left(F_\mathrm{traj}^k \mathbin{||} \Phi_{\mathrm{sem}}(F_\mathrm{sem}^{k})\right)\right)
\end{equation}

During training, ground truth labels $m_{\mathrm{gt}}^k$ are sampled from the ground truth dynamic mask of the first frame based on the 2D tracks $x_1^k$, and the cross-entropy loss is then computed as follows,
\begin{equation}
    \mathcal L=\sum_k -m_{\mathrm{gt}}^k\log(m^k)-(1-m_{\mathrm{gt}}^k)\log(1-m^k)
\end{equation}

Since the 2D tracks are sampled on a grid, during inference, after obtaining the motion confidence scores of trajectories, we apply a dilation operation to generate coarse dynamic masks.


\subsection{SAM-based Post-processing}

Recent dynamic segmentation methods~\cite{zhang2024monst3r,xu2024das3r,chen2025easi3r} typically employ SAMv2~\cite{ravi2024sam2} to refine coarse dynamic masks. These approaches commonly input the entire mask as a query with a single object ID into SAMv2 to generate the final motion masks, which inherently assumes that all masked regions belong to the same object. However, this assumption is often violated in real-world scenarios with multiple dynamic objects, leading to incomplete or even entirely incorrect results.

Inspired by~\cite{tang2025spars3r}, we propose a point-query-based iterative SAM refinement strategy capable of handling multiple dynamic objects in a scene. The detailed procedure is outlined in Algorithm~\ref{algo:sam}. Specifically, we first collect all pixels labeled as dynamic in the coarse mask $M_\mathrm{coarse}$ to form the dynamic pixel set $X_\mathrm{dyn}$. Then, we randomly sample an individual pixel $x$ as a point query for SAM. The resulting SAM-generated mask $M_x$ is then evaluated based on its overlap ratio with $M_\mathrm{coarse}$. If this ratio exceeds a predefined threshold $\beta$ and the number of pixels in $M_x$ meets the minimum requirement $\gamma$, we retain $M_x$ in the final output $M_\mathrm{SAM}$ and remove all its constituent pixels from $X_\mathrm{dyn}$. Otherwise, the sampled query pixel is eliminated from $X_\mathrm{dyn}$. This iterative sampling continues until $X_\mathrm{dyn}$ is empty, after which the union of $M_\mathrm{SAM}$ and $M_\mathrm{coarse}$ is taken as the final dynamic mask $M$.

\begin{algorithm}[]
\caption{SAM-based Post-processing}
\label{algo:sam}
\KwData{RGB Image $I$, Coarse Mask $M_\mathrm{coarse}$, Overlapped Threshold $\beta$, Size Threshold $\gamma$}
\KwResult{Refined Final Dynamic Mask $M$}
SAMSetImage($I$)\;
$X_\mathrm{dyn} \leftarrow$ GetPixels($M_\mathrm{coarse}$)\;
$M_\mathrm{SAM} \leftarrow$ Zeros($M_\mathrm{coarse}$)\;
\While{Size($X_\mathrm{dyn}$) $\neq 0$}{
    $x \leftarrow$ RandomSample($X_\mathrm{dyn}$)\;
    $M_x \leftarrow$ SAMPointQuery($x$)\;
    \If{SumAll($M_x$) $\geq \gamma$ }{
        $M_{\mathrm{overlap}} \leftarrow$ Intersection($M_x,\ M_\mathrm{coarse}$)\;
        $r \leftarrow$ SumAll($M_\mathrm{overlap}$) $\div$ SumAll($M_x$) \;
        \If{$r\geq\beta$}{
            $M_\mathrm{SAM} \leftarrow$ Union($M_\mathrm{SAM}$, $M_x$)\;
            $X_\mathrm{overlap} \leftarrow$ GetPixels($M_\mathrm{overlap}$)\;
            $X_\mathrm{dyn} \leftarrow$ RemovePixels($X_\mathrm{dyn}$, $X_\mathrm{overlap}$)\;
        }
        \lElse {$X_\mathrm{dyn} \leftarrow$ RemovePixels($X_\mathrm{dyn}$, $x$)}
        }
    \lElse {$X_\mathrm{dyn} \leftarrow$ RemovePixels($X_\mathrm{dyn}$, $x$)}
}
$M \leftarrow$ Union($M_\mathrm{SAM}$, $M_\mathrm{coarse}$)
\end{algorithm}

\section{Application to 3D Scene Reconstruction from Dynamic Videos}

To demonstrate the effectiveness of the proposed multimodal dynamic object segmentation, 
we apply it to the problem of static 3D scene reconstruction from dynamic videos. 
In this problem, 
the input video is first converted into multiple video clips of length $N$ using a sliding window approach. Each clip is then processed through a feature fusion network to obtain dynamic masks through feature extraction, trajectory classification and SAM-based post-processing. These dynamic masks are subsequently used to guide staticness-aware 3DGS reconstruction by filtering dynamic elements through static scene attributes. The rendering formulation of staticness-aware 3DGS is defined as follows,
\begin{equation}
    \mathbf c=\sum_{i}c_is_i\alpha'_i\prod_{j=1}^{i-1}(1-s_j\alpha'_j)
\end{equation}
\begin{equation}
    \alpha'_i=\alpha_i\exp(-\frac{1}{2}(x-\mu_i)^T\Sigma_i^{-1}(x-\mu_i))
\end{equation}
where $\mu_i,\ \Sigma_i,\ c_i,\ s_i,\ \alpha_i$ represents the position, covariance matrix, color, staticness, and opacity of the $i$-th 3DGS. The technical details of the 3DGS implementation follow the methodology described in DAS3R~\cite{xu2024das3r}. 
\section{Experimental Results}
\label{sec:experiment}

To validate the proposed method, we conduct extensive experiments on both synthetic and real-world data.

\textbf{Datasets}. We evaluate our approach on the following three  datasets that are widely used in the literature:

\begin{itemize}
\item PointOdyssey~\cite{zheng2023pointodyssey} is a synthetic dataset comprising 131 diverse indoor and outdoor scenes, with approximately 200k images. It includes dynamic objects, camera motion, and provides labels for camera poses and depth, making it well-suited for motion segmentation tasks.

\item DAVIS2017~\cite{perazzi2016benchmark} consists of 90 videos featuring moving objects and cameras, which is relatively closer to the video sequences captured by robot cameras in the real world. Compared to its predecessor, DAVIS2016, this dataset introduces greater complexity by including multiple annotated objects per video, as well as more challenging scenarios involving distractors, occlusions, smaller objects, and intricate structures.

\item Sintel~\cite{butler2012naturalistic} is a synthetic dataset containing 23 sequences, renowned for its dynamic complexity. It presents highly challenging scenes with detailed motion patterns and interactions between dynamic and static elements, making it a valuable resource for evaluating motion segmentation methods.
\end{itemize}

\textbf{Metrics}. For dynamic object segmentation, we report accuracy, IoU, precision and recall metrics by comparing the predicted and ground-truth masks. For static scene reconstruction from dynamic videos, we report the PSNR and SSIM metrics by comparing the predicted and ground-truth RGB images.

\textbf{Baselines}. We benchmark our approach against the SOTA methods including the optical flow-based ParticleSfM (P-SfM)~\cite{zhao2022particlesfm}, end-to-end reconstruction based methods MonST3R~\cite{zhang2024monst3r} and Easi3R~\cite{chen2025easi3r}, and 3DGS-based DAS3R~\cite{xu2024das3r}. We use the official open-source implementations of these methods to evaluate the quality of motion masks. For reconstruction quality, we adopt the static background reconstruction with 3DGS in DAS3R and integrated the motion masks produced by different methods into it for comparison. For our methods, we use \textbf{Ours} to refer without any post-processing, \textbf{Ours(+SAM)} to refer the proposed point-query-based iterative SAM refinement strategy.

\textbf{Implementation Details}. We take video clips with $N=10$ frames as input. In the main model, the Transformer we use consists of two encoder layers and two decoder layers, and the architecture of the aggregation networks we used based on  OANet~\cite{zhang2019learning} with modifications. For the hyperparameters mentioned in the paper, we set $\alpha=0.7$, $\beta=0.3$, and $\gamma=5$. We use the AdamW optimizer with a maximum learning rate of $0.001$ and a batch size of 4 per GPU. The training stage is performed on $2\times$ A100-80G GPUs for 50 epochs with 1000 steps per epoch, using only the training split of the PointOdyssey dataset. 

\subsection{Motion Mask Accuracy}

\begin{table}[ht]
\setlength{\tabcolsep}{1.3mm}
\renewcommand\arraystretch{1.1}
\caption{Motion Mask Accuracy of PointOdyssey dataset. \textbf{Acc} for accuracy, \textbf{Pre} for precision and \textbf{Rec} for recall in percentage.}
\vspace{-2mm}
\label{tab:mask-po}
\hspace*{0cm}\makebox[\linewidth][c]{%
\begin{tabular}{ c | c | c | c | c }
\toprule
Methods & Acc (mean) & IoU (mean) & Pre (mean) & Rec (mean)  \\
\hline
P-SfM & 88.50 & 27.39 & 56.59 & 34.99 \\
MonST3R & 85.27 & 18.61 & 59.59 & 22.11 \\
Easi3R & 88.17 & 40.01 & 59.45 & 57.49 \\
DAS3R & {\bf \color{bronze}92.22} & {\bf \color{gold}66.54} & {\bf \color{gold}76.62} & {\bf \color{bronze}79.64} \\
Ours & {\bf \color{gold}93.70} & {\bf \color{bronze}63.37} & {\bf \color{bronze}73.45} & {\bf \color{silver}81.42} \\
Ours(+SAM) & {\bf \color{silver}93.69} & {\bf \color{silver}64.38} & {\bf \color{silver}73.79} & {\bf \color{gold}82.23}\\
\bottomrule
\end{tabular}}
\vspace{-1mm}
\end{table}

\begin{figure*}
    \centering
    \includegraphics[width=\linewidth]{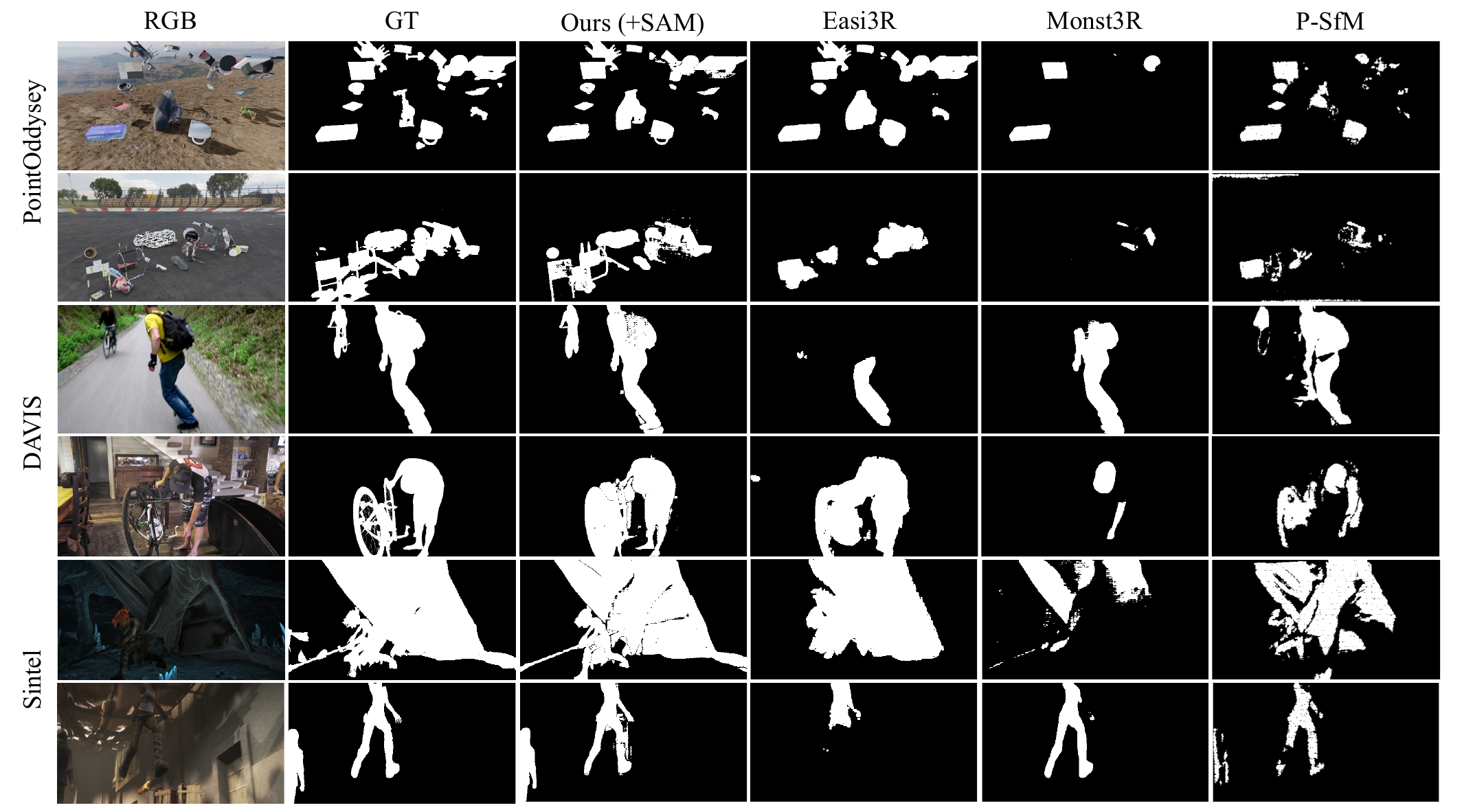}
    \vspace{-5mm}
    \caption{\textbf{Visual Comparison of Dynamic Mask Prediction.} We provide quantitative comparisons for dynamic mask prediction tasks. Our method produces more precise and complete masks compared to other approaches.}
    \label{fig:motion_mask}
\end{figure*}

\subsubsection{Comparison on PointOdyssey} Detailed comparison results are shown in Tab.~\ref{tab:mask-po}. Except for ParticleSfM, all other methods were trained on this dataset. Our approach demonstrates superior performance in terms of accuracy and recall. Compared to ParticleSfM, which primarily relies on optical flow, and MonST3R, which mainly depends on depth estimation, our method effectively integrates 2D and 3D information, enabling it to better handle scenarios where both objects and the camera are in motion. Additionally, our approach remains robust even when one type of feature degrades, ensuring consistent motion recognition results. 

Easi3R, which extracts attention layer features from MonST3R, suffers from accuracy limitations due to its dependence on implicit reconstruction precision. Furthermore, its lower resolution results in less accurate object contours. In contrast, we leverage SAM to extract more explicit semantic information and utilize feature clustering to achieve more comprehensive identification of dynamic objects. Combined with post-processing, our approach further enhances the refinement of object contours.

DAS3R builds on MonST3R by adding a motion mask head and training with pre-trained models using 3DGS. However, our method achieves superior results with significantly fewer training steps, demonstrating greater efficiency and effectiveness.

\begin{figure*}
    \centering
    \includegraphics[width=0.9\linewidth]{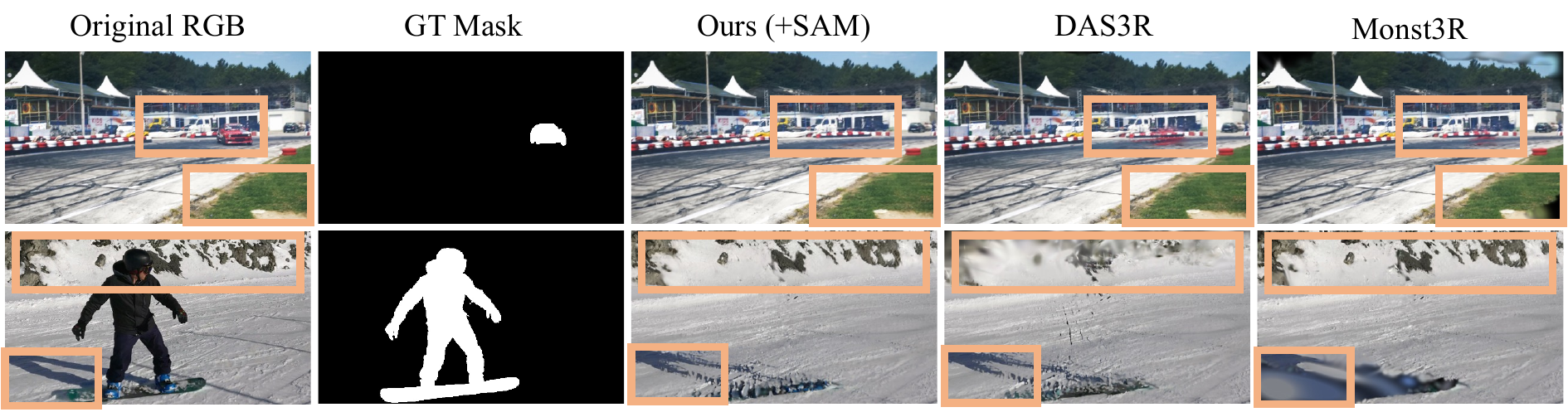}
    \vspace{-5mm}
    \caption{\textbf{Visual Comparison of Static Scene Reconstruction.} We provide quantitative comparisons for the tasks of static scene reconstruction from dynamic videos using the DAVIS2017 dataset. Our method achieves more photorealistic novel view synthesis results compared to other methods, demonstrating its effectiveness in dynamic object segmentation.}
    \label{fig:recon}
    \vspace{-3mm}
\end{figure*}

\begin{table}[ht]
\setlength{\tabcolsep}{1.3mm}
\renewcommand\arraystretch{1.1}
\caption{Motion Mask Accuracy of DAVIS2017 dataset. \textbf{Acc} for accuracy, \textbf{Pre} for precision and \textbf{Rec} for recall in percentage.}
\vspace{-2mm}
\label{tab:mask-davis}
\hspace*{0cm}\makebox[\linewidth][c]{%
\begin{tabular}{ c | c | c | c | c }
\toprule
Methods & Acc (mean) & IoU (mean) & Pre (mean) & Rec (mean)  \\
\hline
P-SfM & {\bf \color{bronze} 90.21} & 37.21 & 58.87 & 53.65 \\
MonST3R & 87.77 & 41.57 & 58.45 & 50.47 \\
Easi3R & 88.64 & {\bf \color{bronze} 50.69} & {\bf \color{silver}61.25} & 78.03 \\
DAS3R & 86.96 & 44.83 & 48.79 & {\bf \color{bronze} 83.01} \\
Ours & {\bf \color{silver}92.89} & {\bf \color{silver}53.67} & {\bf \color{bronze}59.72} & {\bf \color{silver}86.76} \\
Ours(+SAM) & {\bf \color{gold}94.01} & {\bf \color{gold}59.77} & {\bf \color{gold}62.22} & {\bf \color{gold}93.14}\\
\bottomrule
\end{tabular}}
\vspace{-1mm}
\end{table}

\subsubsection{Comparison on DAVIS} The comparison results of DAVIS2017 are shown in Tab.~\ref{tab:mask-davis}. DAVIS2017 is a dataset composed of real-world scenes, where our method consistently outperforms others across all metrics, demonstrating excellent generalization capabilities. By converting absolute information such as pixel positions, depth, and poses into relative representations like optical flow, scene flow, and camera motion, and representing them using point trajectories, our approach effectively reduces overfitting to specific scenes. Qualitative visualizations of the motion masks are shown in Fig.~\ref{fig:motion_mask}.

\begin{table}[ht]
\setlength{\tabcolsep}{1.3mm}
\renewcommand\arraystretch{1.1}
\caption{Motion Mask Accuracy of Sintel dataset. \textbf{Acc} for accuracy, \textbf{Pre} for precision and \textbf{Rec} for recall in percentage.}
\vspace{-2mm}
\label{tab:mask-sintel}
\hspace*{0cm}\makebox[\linewidth][c]{%
\begin{tabular}{ c | c | c | c | c }
\toprule
Methods & Acc (mean) & IoU (mean) & Pre (mean) & Rec (mean)  \\
\hline
P-SfM & 79.55 & 27.10 & 59.69 & 31.76 \\
MonST3R & 72.97 & 30.18 & 53.34 & 34.57 \\
Easi3R & 80.96 & 37.51 & {\bf \color{bronze}62.98} & 48.44 \\
DAS3R & {\bf \color{bronze}81.06} & {\bf \color{silver}53.36} & 58.17 & {\bf \color{gold}82.63} \\
Ours & {\bf \color{silver}87.33} & {\bf \color{bronze}46.04} & {\bf \color{silver}65.99} & {\bf \color{bronze}57.11} \\
Ours(+SAM) & {\bf \color{gold}89.99} & {\bf \color{gold}54.66} & {\bf \color{gold}67.43} & {\bf \color{silver}67.97} \\
\bottomrule
\end{tabular}}
\vspace{-1mm}
\end{table}

\subsubsection{Comparison on Sintel} 
The accuracy of motion mask estimation is compared in Tab.~\ref{tab:mask-sintel}. On the Sintel dataset, the performance of all methods is suboptimal. While DAS3R achieves relatively high recall, its other metrics are notably low, suggesting a trade-off between quality for quantity. In contrast, our method demonstrates a more balanced performance across metrics. The MPI Sintel dataset is characterized by simple camera movements, unrealistic appearances, generally textureless scenes, and minimal 3D structure once dynamic objects are removed. These factors make scene reconstruction highly prone to failure. Moreover, the dataset frequently contains images dominated by dynamic objects, which presents significant challenges for dynamic object segmentation.

\begin{figure*}
    \centering
    \includegraphics[width=0.8\linewidth]{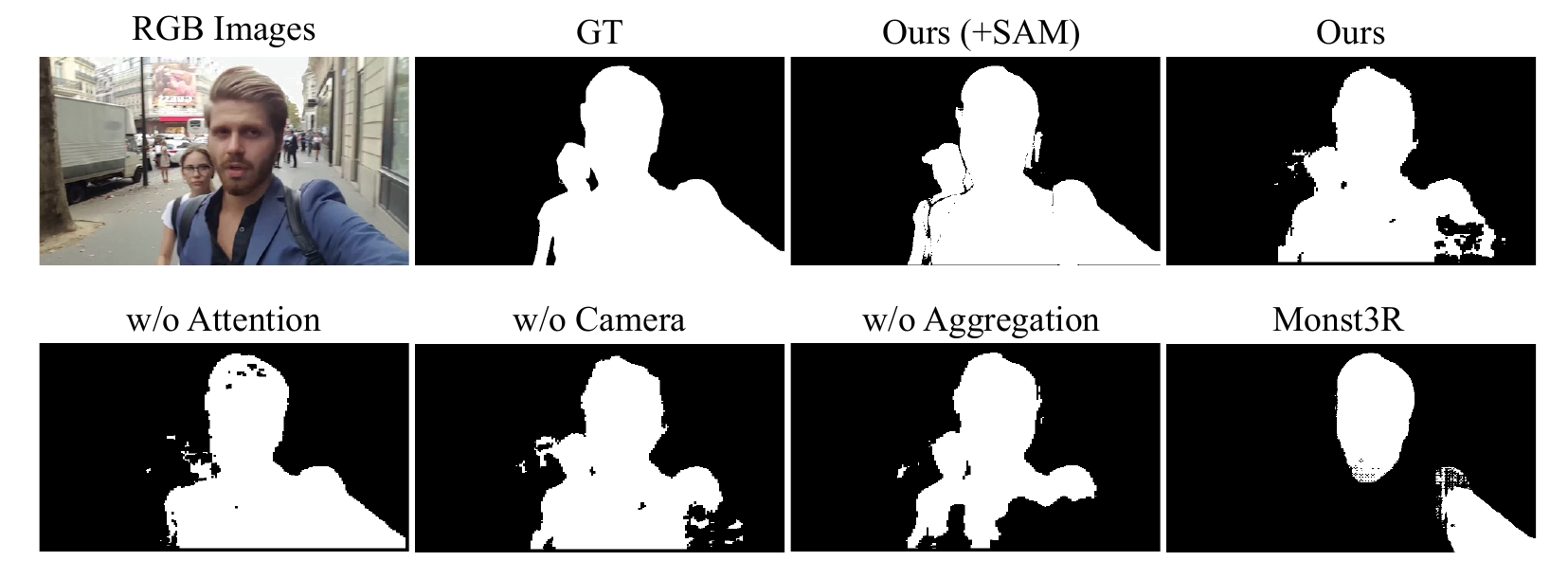}
    \vspace{-2mm}
    \caption{\textbf{Visual Comparison in the Ablation Study.} We conduct ablation study on different feature combinations and processing strategies. Our full model achieves the best performance among all variants. }
    \label{fig:ablation}
    \vspace{-2mm}
\end{figure*}

\begin{figure*}
    \centering
    \includegraphics[width=0.8\linewidth]{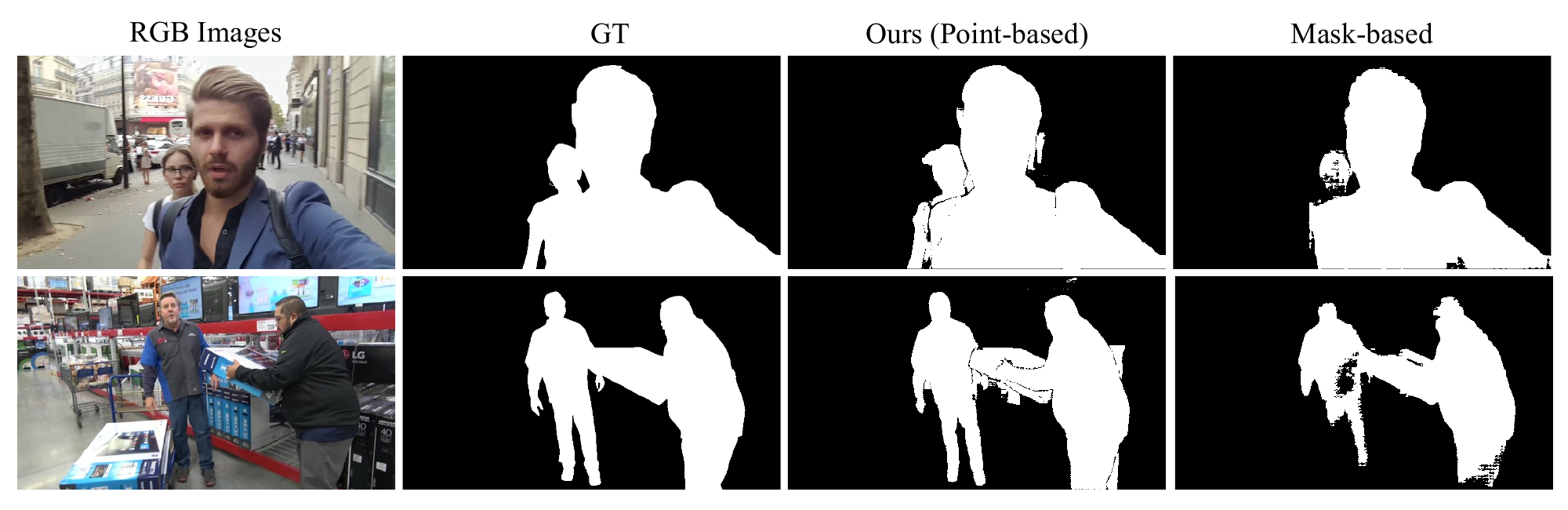}
    \vspace{-2mm}
    \caption{\textbf{Comparison of Different SAM refinement strategy.} We show comparisons between ours and mask-based SAM refinement strategy. Previous mask-based refinement strategy suffers from incomplete mask generation due to the assumption that dynamic pixels belong to a single object.}
    \label{fig:ablation_sam}
    \vspace{-5mm}
\end{figure*}

\subsection{Static Scene Reconstruction}
\begin{table}[ht]
\renewcommand\arraystretch{1.1}
\caption{Static scene reconstruction quality of DAVIS2017 dataset. The DAS3R split includes: blackswan, camel, car-shadow, dog, horsejump-high, motocross-jump, parkour, soapbox}
\vspace{-2mm}
\label{tab:rec-davis}
\hspace*{0cm}\makebox[\linewidth][c]{%
\begin{tabular}{ c | c c | c c  }
\toprule
\multirow{2}{*}{Methods} & \multicolumn{2}{c | }{DAS3R split} & \multicolumn{2}{c  }{All}   \\
 & PSNR & SSIM & PSNR & SSIM \\ 
\hline
MonST3R & 27.19 & 0.863 & 26.88 & 0.874 \\
Easi3R & 29.14 & 0.905 & 27.53 & 0.882 \\
DAS3R & 28.40 & 0.896 & 27.28 & 0.871 \\
Ours(+SAM) & {\bf \color{gold}30.60} & {\bf \color{gold}0.933} & {\bf \color{gold}28.90} & {\bf \color{gold}0.903} \\
\bottomrule
\end{tabular}}
\vspace{-3mm}
\end{table}

We follow the reconstruction pipeline of DAS3R to obtain camera poses, depth, motion masks, and other relevant information, which are then used as inputs to train a 3DGS model for static scenes. Using the camera poses from the training views, we render images without dynamic objects. The DAVIS dataset provides ground-truth motion masks, which we use as a baseline. To evaluate the quality of static scene reconstruction, we compare the rendered images obtained by training 3DGS with motion masks from other methods against those rendered using ground truth motion masks. Detailed quantitative comparison results are shown in Tab.~\ref{tab:rec-davis}. Our method achieves highest PSNR and SSIM on both the DAS3R split and all scenes of the DAVIS2017 dataset. We also provide qualitative comparisons in Fig.~\ref{fig:recon}. The rendered images generated by our method exhibit fewer artifacts and more accurately capture and remove dynamic objects. Furthermore, the background is rendered with high clarity, indicating that the dynamic foreground and static background have been precisely and effectively separated.

\subsection{Ablation Study}

\begin{table}[ht]
\renewcommand\arraystretch{1.1}
\caption{Motion Mask Accuracy of DAVIS2017 dataset with different feature combinations and processing strategy.}
\vspace{-2mm}
\label{tab:ablation-davis}
\hspace*{0cm}\makebox[\linewidth][c]{%
\begin{tabular}{ c | c | c | c | c | c }
\toprule
\multicolumn{4}{c | }{Features} & Acc  & Rec  \\
Camera & Attention & Aggregation & SAM & mean & mean \\
\hline
\ding{56} & \ding{52} & \ding{52} & \ding{56} & 92.11 & 84.22 \\
\ding{52} & \ding{56} & \ding{52} & \ding{56} & 87.67 & 68.20 \\
\ding{52} & \ding{52} & \ding{56} & \ding{56} & 92.51 & 84.46 \\
\ding{52} & \ding{52} & \ding{52} & \ding{56} & 92.89 & 86.76 \\
\ding{52} & \ding{52} & \ding{52} & \ding{52} & 94.01 & 93.14 \\
\bottomrule
\end{tabular}}
\end{table}

We evaluate our method by testing different feature combinations on the motion mask accuracy, i.e., camera pose embeddings, attention features similar to Easi3R, feature aggregation modules and SAM refinements. Tab.~\ref{tab:ablation-davis} and Fig.~\ref{fig:ablation} report the results tested on DAVIS2017. Attention features are directly extracted from MonST3R layers and camera poses are further optimized by bundle adjustment based on depths and confidences inferred from MonST3R. Although pose information is inherently embedded in optical flow and depth, explicitly providing poses as input features to the model still proves beneficial. This simplifies the learning process for the model and allows for more direct and effective utilization of camera motion. 
After clustering and aggregating the point track features with our aggregation module, the model can adaptively identify which type of feature should dominate based on the characteristics of each scene, while also mitigating the impact of feature degradation, such as imprecise optical flow and depth estimation. Moreover, SAM refinement can help the predicted mask better aligned with the object boundaries, resulting in a higher recall rate.

We also provide a comparison of different SAM refinement strategies, as shown in Fig.~\ref{fig:ablation_sam}. It is evident that without SAM refinement, the predicted masks are incomplete, resulting in a lower recall rate. If the mask-based SAM refinement is applied, due to the assumption that all masked pixels belong to a single object, this strategy may not be suitable for all cases in the real world, and the output predicted masks may suffer from poor shapes.

\subsection{Efficiency Analysis}

We present the model sizes and inference time of our method and baseline models in Tab.~\ref{tab:efficiency}. Compared with state-of-the-art (SOTA) methods, our approach features more compact model size and shorter inference time. We also provide the average inference times of the mask-based SAM refinement strategy used in previous works and our point-based refinement strategy. It is worth noting that the number of loops in our method is directly determined by the number of masked pixels in the coarse dynamic mask, therefore, the inference time may vary depending on the input video clips.

\begin{table}[ht]
\renewcommand\arraystretch{1.1}
\caption{Model Efficiency Analysis.}
\vspace{-2mm}
\label{tab:efficiency}
\hspace*{0cm}\makebox[\linewidth][c]{%
\begin{tabular}{ c | c | c}
\toprule
Model & Params (M) & Inference Time (s) \\
\hline
P-SfM & 0.5377 & 14.0724 \\
MonST3R & 571.171 & 87.0266 \\
Easi3R & 571.171 & 37.0712 \\
DAS3R & 611.868 & 27.5518 \\
Ours & 2.798 & 0.2833 \\
Mask-based SAM Refine & - & 2.6565 \\
Ours (Full Sequence) & - & 32.4399 \\
\bottomrule
\end{tabular}}
\end{table}
\section{Conclusions and Future Work}

We propose a novel dynamic object segmentation framework by unifying multimodal cues, including 2D point tracks, 3D reconstruction results, and semantic features. This framework enables the model to adaptively determine which type of feature should dominate based on the characteristics of each scene, while also mitigating the impact of feature degradation. Moreover, we introduce a novel point-query-based SAM post-processing method capable of handling multiple objects within a single mask. Extensive experiments demonstrate the effectiveness of our model. However, despite strong performance on various datasets, our method occasionally encounters challenges in scenarios where dynamic objects dominate the image. Addressing this limitation through more diverse training data and enhanced model refinements will be a key focus of our future work.

\textbf{Acknowledgement}. This work is supported by the Shenzhen Science andTechnology Program under Grant Nos. KJZD20230923115210021.




\footnotesize


\bibliographystyle{IEEEtran} 
\bibliography{icra_abrv}

\end{document}